# Finding Inner Outliers in High Dimensional Space

Zhana

**ABSTRACT**

Outlier detection in a large-scale database is a significant and complex issue in knowledge discovering field. As the data distributions are obscure and uncertain in high dimensional space, most existing solutions try to solve the issue taking into account the two intuitive points: first, outliers are extremely far away from other points in high dimensional space; second, outliers are detected obviously different in projected-dimensional subspaces. However, for a complicated case that outliers are hidden inside the normal points in all dimensions, existing detection methods fail to find such inner outliers. In this paper, we propose a method with twice dimension-projections, which integrates primary subspace outlier detection and secondary point-projection between subspaces, and sums up the multiple weight values for each point. The points are computed with local density ratio separately in twice-projected dimensions. After the process, outliers are those points scoring the largest values of weight. The proposed method succeeds to find all inner outliers on the synthetic test datasets with the dimension varying from 100 to 10000. The experimental results also show that the proposed algorithm can work in low dimensional space and can achieve perfect performance in high dimensional space. As for this reason, our proposed approach has considerable potential to apply it in multimedia applications helping to process images or video with large-scale attributes.

## Categories and Subject Descriptors
H.2.8 [**Database Management**]: Database Applications−*Data mining*; H.3.4 [**Information Systems**]: System and Software−*Performance evaluation (efficiency and effectiveness)*

## General Terms
Algorithms, Experimentation, Performance

## Keywords
Outlier Detection, High Dimensional Space, Projected Dimension, PCD

## 1. INTRODUCTION
Outlier detection is an essential part of data mining tasks, which has many practical applications in different domains, such as fraud detection, medicine development, public health management, sports statistics analysis, etc. Outlier detection in high dimensional space is even more significant since the ever-increasing data emerged from the advent of the Internet and the social networks in last decades. Therefore, how to detect outliers explicitly in the large dataset and how to discover them with an effective method become a more urgent issue. Though many researchers have mentioned various outlier definitions, the most cited definition is Hawkins': an outlier is an observation that deviates so much from other observations as to arouse suspicion that it generated by a different mechanism [1]. This definition not only describes the outliers in an observing or measuring way but also points out the outliers' essential difference by the generation mechanism. According to the different generation mechanisms between normal points and outliers, outliers may be defined as a different distribution from normal points' distribution within the same ranges. In this paper, we define the outliers and normal points conforming to random distribution and normal distribution separately. The outliers are located in the center part of normal range in all dimensions, and they seem to be inner points. That is we call them "inner outliers".

The ever-proposed outlier detection approaches in high dimensional space have faced the serious issue, i.e. "Curse of dimensionality". Two categories of solutions are proposed to improve them: one is to insist former distance methods with more robust function to find outliers in full-dimensional space; the other is to find outliers in some certain projected dimensional subspaces using grid or cell of the original feature space. The first solution includes Hilout[4], LOCI[6], GridLOF[19], ABOD[5], etc. The second solution includes Aggarwal's Fraction[3], GLS-SOD[10], CURIO[9], SPOT[13], Grid-Clustering[19], etc. The second one has been accepted widely as a prime candidate solution in high dimensional space for the better accuracy and adaptive performance. However, ever-proposed algorithm cannot work well in some cases, i.e. not all outliers can be found just using low projected-dimensional subspace. Although some points seem no different with normal points in low projected-dimensional subspace, they may be real outliers in high dimensional space. Therefore, a more robust and precise approach to improve the subspace detection still needs to be considered.

In this paper, we propose a robust subspace detection algorithm PCD (Projected Cell Detection), which utilizes twice dimension projections. The first projection employs a local density calculating method in projected dimensional subspace with the ratio against average density. The second projection maps the points in the grid cells from original projected-dimension to other projected dimensions to evaluate these points. After that, each point's average weight values in above two steps are calculated. The points scoring the largest values of weight are taken as outliers.

The features and achievements of our proposal are summarized as follows:

- The subspace based outlier detection solutions is improved with twice dimension projection by our proposal, which introduce m weights to each point in the first projected-dimensional subspace and also introduce m×(m-1) weights to each point in the second projected-dimensional space. By these twice projections, our proposal can detect the outliers that cannot be found effectively in low-projected dimensional subspaces by ever proposed subspace detection methods.

- Cells that contain the least points are taken as outliers inside in the ever-proposed methods, while cell values related to the points are considered to detect outliers in our proposal that improve the outlier detection. Each point is denoted by the cell values in every projected dimension. The points scoring the highest weight are true outlier in statistic probability.

- Based on the experiment, high dimensional datasets that are generated to the Hawkins' outlier definition, it has been found that our proposal is more stable and precise in large volumes of data in high dimensional space, compared with ever-proposed outlier detection algorithms.

- In our proposal, every calculation process is restricted in no more than two dimensions, so it avoids calculation in the high dimensional space. Therefore, the "curse of dimensionality" problem does not happen in this case. Unlike ever-proposed algorithms such as the dimension reduction and the subspace detection, the data information do not lose in our algorithm.

The rest of the paper is organized as follows: Section 2 gives a brief overview of related works on high dimensional outlier detection. Section 3 gives necessary definitions and equations, and then introduces the algorithm. Section 4 evaluates the proposed method by experiments of different dimensional datasets. Finally, we conclude in section 5.

## 2. RELATED WORKS

For last ten years, many studies have been conducted on outlier detection with large datasets that can be categorized into the following four groups.

The methods in the first group are based on the measurement of distance or density as used in the traditional low dimensional outlier detections. The most used method is based on local density-LOF[2]. It uses the MinPts-nearest neighbor concept to detect outliers. The normal points' values are approximately 1 in this method. Hence, the points whose values are obviously larger than 1 or some top largest-value points are outliers. The LOF works well in a low dimensional dataset and is partly practical in a high dimensional dataset.

The methods in the second group are based on the subspace clustering method. Some high dimensional outliers also seem deviated from others in low dimensional space. The outlier points can be regarded as byproducts of clustering methods. The Aggarwal's method [3] belongs to this group. He uses the equi-depth ranges in each dimension, with expected fraction and deviation of points in k-dimensional cube D given by $N \times f^k$ and $\sqrt{N \times f^k \times (1-f^k)}$. This method detects outliers by calculating the sparsity of coefficient $S(D)$.

The methods in the third group are the outlier detection with dimension deduction, such as SOM (Self-Organizing Map) [17, 18], mapping several dimensions to two dimensions, and then detecting outliers in two dimensional space. This method may cause information lost while the dimension deducts. Thus, it is only used in some special dataset.

The forth group includes other methods such as ABOD [5] and RIC[7, 8]. ABOD(Angel-Based Outlier Detection) is an algorithm with globe method, and it is based on the angle concept of vector product and scalar product. The outliers have the small angles since they are far from other points. Another algorithm in this group is called RIC (Robust Information-theoretic Clustering), which uses integer-coding points in each dimension, then detects cluster with MDL (Minimum Description Length).

From above mentioned, it can be said that the most measurements on outlier detection are based on distance or density between points. The good aspect is that the concepts is straightforward to image and easy to realize. The bad aspect is the occurrence of "curse of dimensionality". Almost these methods work smoothly in the relatively low dimensional space, and the result can be shown clearly. However, the problem is raised in extremely high dimensional datasets. As the number of dimensions in a dataset increases, the distance measures become increasingly meaningless because the points spread out in extremely high dimensional space and they are almost equal in distance each other. Most methods try to mitigate the curse-of-dimensionality effects by subspace clustering. However, the main trap into which we fall, is that blindly apply concepts to high dimensional space that are valid in two or three -dimensional space. One example is the use of the Euclidean distance that behaves poorly in high dimensional space. Another example is misjudged outliers, which are detected in low dimensional space, as all outliers existed in high dimensional space. Hence, we need to find a new method to detect outliers in a different view in order to avoid the "curse of dimensionality".

## 3. PROPOSED METHOD
### 3.1 General Idea

For the purpose of data analysis in an intuitive way, the dataset is always mapped to a set of points in a certain dimensional space. Each observation is considered as a data point, and each attribute associated with it is considered as a dimension. Thus, the set of attributes of the dataset constitutes a dimensional space. Our method also analyzes data in the point's space.

After reviewing the ever-proposed methods of subspace detection, we have found a question by experiments that points detected as outliers in all subspaces are real outliers even in high dimensional space, but some points neglected in all subspaces may also be outliers if observing them in high dimensional spaces. In addition, the accuracy of those methods is even worse with partly random subspace detection. Therefore, more robust subspace detection approach needs to be studied.

Our proposed approach solves this issue in two aspects. First, we collect the point's value from relevant local density value in each projected dimensions instead of only calculating grid cell value as seen in ever-proposed methods. The cell value denotes the weight of the related point in each projected dimensions. The multiple weights of each point in all dimensions guarantee the accuracy of outlier detection that points to be evaluated in a statistic way. Second, we employ a twice dimension-projection method, which detects outliers in one-dimensional subspace (the first dimension-projection), and maps a certain region points to other projected-dimension from the original projected dimension (second dimension projection) to measure the points deviation. The points deviated from other points in a new projected dimensions are most possible outliers in high dimensional space. These inner outliers hidden in a normal points' distribution cannot be detected by ever-proposed subspace detection approaches. While in our proposal, the inner outliers are detected successfully by comparing points, which scoring the top weight values in the above two steps of projection are judged.

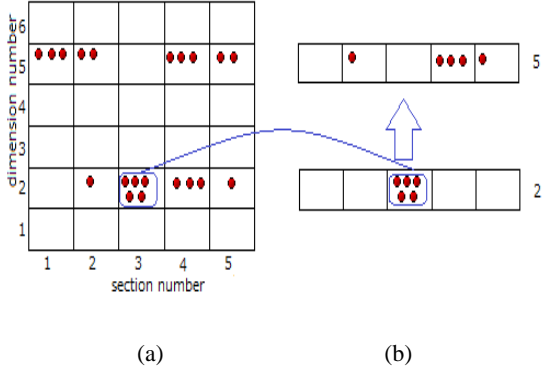

(a)  (b)

**Figure 1: Space Division and Dimension Projection**

The dimensions are mutually independent, and the data distributions vary in different dimensions. Hence, it is difficult to compare all points' distributions in general data space. In this paper, we take the points of a small region, which is called a cell, in one projected dimension for analysis, and evaluate their distribution changes in other projected dimensions. For this reason, the data ranges are divided into the same number of equal width cells in each dimension, which means points in the cell have the equal probability. A cell contains co-related data points. Therefore, the data space looks like a big cube with lots of small cubes inside in the case of three-dimensional space. We utilize the points' local density in a cell to denote the point weight value during the twice dimension-projection process. The new data space is similar to a grid cell as shown in Figure 1(a). In this example, the ten points have different distributions in dimension 2 and dimension 5. We observe the distribution change of five points' in the $3^{rd}$ cell of dimension 2 when they are projected to dimension 5. These points may distribute scattered in different cells in dimension 5 as shown in the Figure 1(b). In this example, after the projecting points in the cell 3 from dimension 2 to dimension 5, we can find that a point in the $2^{nd}$ cell of dimension 5 is far from other points in the $4^{th}$ and the $5^{th}$ cells. In this case, the hidden outlier in dimension 2 can be detected.

In our proposal, the conventional relationship between points and dimensions are replaced with the new relationship between points, dimensions and cells as shown in Figure 2. The point's weight value is decided by the cell value in a certain projected-dimension. The cells containing the number of points in one dimension are used to compute point's density ratio. The mapping process from one projected dimension to a new projected one is based on the information of the points and their cells in the two projected dimensions. Furthermore, the original data are decomposed into the point information and the cell information. The point's weight is also determined by computing the cell information where the points are located. The cell that is used for calculation depends on the point in different projected dimensions.

The three steps of PCD algorithm structure are also shown in Figure 2. In the first step, it calculates *CellVal* in one-projected dimensional subspace. In the second step, it calculates *CellValp* in a second projected dimensional subspace. In the last step, it collects all cell values of co-related points denoting point values, and finally it sums up all point value with a statistic approach. Then, the points with largest weights are taken as outliers.

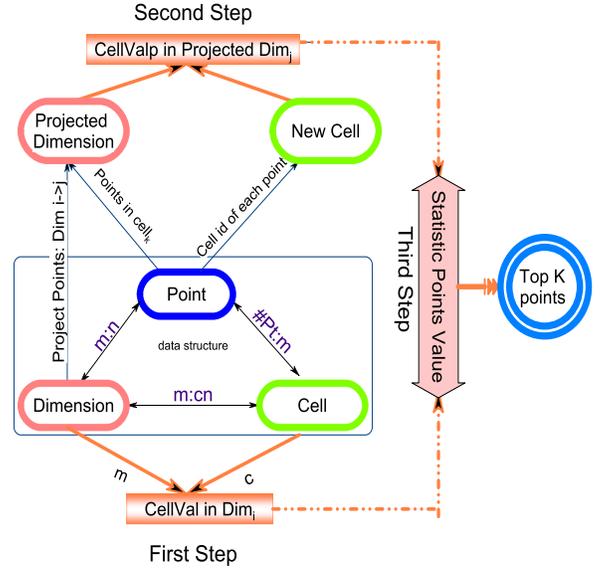

**Figure 2: PCD Algorithm Structure and Points-Dimensions-Cell Relationship**

## 3.2 Definitions and Data Structures

In this section, we present the new definitions of outlier used for calculation in our proposal, and we introduce necessary definitions to describe our outlier detection algorithm.

Let DB be an *m* dimensional dataset including *n* points. Each dimensional range is divided into *c* cells.

**DEFINITION 1** (*Outlier*)
Let *p* be a point of **DB**, then the weight of *p* in **DB** is defined as a serious of weights in first projected dimensional subspace and second projected dimensional subspaces. The weights of cells denote the corresponding point's weights. Then, *p* is the $k^{th}$ outlier in DB, denoted as $outlier^k$, if there are exactly *k-1* points *q* in **DB** such that weight $w(q) \geq w(p)$. We denote with $Out^k$ the set of the top *k* outliers of **DB**.

The detailed calculation of weight of point p is introduced in section 3.3 with three equations. The other useful definitions are a list in Table 1.

**Table 1. List of Definitions and Symbols**

| |
|---|
| ***p***: the information of point. $p_j$ refers to the $j^{th}$ point of all points. $p_{i,j}$ refers to the $j^{th}$ point in $i^{th}$ dimension. |
| ***Cell***: the range of data in each dimension is divided into the same number of equal-width fractions, which is called a cell. |
| ***cn***: the cell number in each dimension. As shown in the grid structure in Figure 1, *cn* is defined equally in all dimensions. |
| ***Density***: count of points in one cell is called *density* or *cell density*. |
| ***Cell-cluster***: the continuous cells which have points in are called a cell-cluster in second projected dimensional subspace. |
| ***CluLen***: the length of a cell cluster. It is measured by the |

number of cells in the cell cluster. See Equation (2).

***CellVal***: the cell value which is calculated by density ratio in projected dimensions. *CellVal* denotes *PtVal*, which is the point's weight value in the cell. The detail of *CellVal* and *PtVal* is defined in Equation (1) and (2).

***SI***: the statistic information of each point denotes the point's weight, as defined in Equation (3).

The cell in our algorithm is similar to the ever-proposed grid cell. The difference is that the points of a cell are not only calculated for density ratio, but also used for dimension projection.

The *CellVal* and *CellValp* are cell values with different definitions. The similarity of the *CellVal* and *CellValp* is that they are calculated cell values instead of point, and they denote the corresponding point with proper weight value. The difference is that they are calculated with different equations in different projected dimensions. The *CellVal* is obtained by the calculation of point's cell density in the first projected-dimension, while the *CellValp* is obtained by the calculation of point's cell scattered in the second projected dimensions. The number of *CellVal* of the point is *m*, while the number of *CellValp* is $m \times (m-1)$ at most. The further details about *CellVal* and Cell*Valp* are also discussed in the next section.

In our model, the *cn* is identical in each projected dimension, and it is only related to the total number of points and the average value of *cell density*. Since the *cn* ensures the same number of cells in all dimensions, the *cell density* can reflect the points' distributions in different dimensions. The outliers are easily detected by comparing the cell density in different projected dimensions.

If outliers have low density in some low dimensions, outliers can be detected by any subspace detection methods. However, they do not always appear like that. Therefore, the points need to be further detected in second projected dimensions. In our proposal, the points of a cell are projected to other dimension iteratively to observe the different distribution in that dimension. The densities are recalculated in that dimension, and the cell cluster is used to measure their dispersion.

### 3.3 Projected Cell Density

Distinguished from the ever-proposed methods whose calculations focus on the points, all the calculations are based on the cells instead of points in our proposal. Then, points become a bridge connecting different dimensions. Therefore, hereafter we call the proposed method "Projected Cell Density Method", or PCD in short.

We divide our proposal into three steps. The first step is calculating *CellVal($p_{i,j}$)* by the ratio of *cell density* against average in $i^{th}$ dimension, and get all *CellVal(p)* in all dimensions by the loop of dimensions. The second step gets *CellValp($p_{i,j}$,k)* with *cell-cluster* density ratio in $k^{th}$ projected dimensions from the $i^{th}$ dimension and gets all *CellValp(p)* by the loop of projected dimensions. In the last step, we sum up all weight values of the points and get statistic information for each point by Equation (3). So, the *SI($p_i$)* is calculated from *CellVal($p_i$)* in first projected dimensions and *CellValp($p_i$)* in second projected dimensions.

**Step 1: Calculate cell value denoting the point value in one-projected dimensional subspace**

Detecting outliers in each dimension is a relatively familiar task since it is similar to other algorithms effective in low dimensional subspace. The calculation of density ratio is better than the difference of densities, because it need not consider stand deviation for its distribution. By the ratio against average value, we can know which cell contains the least points in each dimension. *CellVal* donates the related *PtVal* of point in the dimension. Summing up all *CellVal* by the loop of dimensions, the points with the least weights are outliers. The calculation formula of the cell density ratio against average cell density is defined in Equation (1), w.r.t. a point $p_{i,j}$ in the $Cell_{i,k}$ in the $i^{th}$ dimension.

$$PtVal_{i,j} \Leftarrow CellVal(p_{i,j}) = \frac{density(Cell_{i,k})}{\frac{density(Cell_i)}{Card(Cell_i)}} \quad (1)$$

Where *density($Cell_{i,k}$)* refers to the $k^{th}$ cell density of the $i^{th}$ dimension, and $p_{i,j}$ is the $j^{th}$ point in $i^{th}$ dimension, which is in the $k^{th}$ cell. The average cell density in $i^{th}$ dimension is obtained from *density($Cell_i$)* divided by *Card($Cell_i$)*, where *density($Cell_i$)* means the total value of *cell density* and *Card($Cell_i$)* means the number of cells.

By Equation (1), the point's *PtVal* gets the value from the *CellVal* in each dimension and C*ellVal* denote the points in the cell with proper *PtVal*. After calculation with the dimension loop, the points with the least total *PtVal* are more probably to be outliers.

For example, taking the Figure 1 to consider the 5 points of the 3$^{rd}$ cell in dimension 2, the *CellVal* of each point is C*ellVal($P_{2,j}$)=2.5 (j=1,2,3,4,5)*. It means the 5 points have the same *CellVal*.

The *density(Cell)* is often different in each dimension. We have considered the particular case that most of the points are only placed in several cells and no point in other cells in some certain dimensional space. In this case, the average *cell density* becomes very low when all the cells are counted with points and without points, and the outliers get higher cell density ratio in these dimensions. This high cell density ratio may make the outlier detection more difficult in these dimensions and even cannot distinguish the outliers in the worst case. Hence, we neglect the cells without points, and only count the cells with points when calculating *density($Cell_i$)*.

**Step 2: Calculate cell-cluster value denoting point value in second projected dimensional space**

If outliers are hidden behind normal points in some dimensions, they seem normal, too. These outliers are occasionally mixed with normal clustered points in different dimensions. The subspace detection methods are not able to find such outliers under this situation. They are only found as normal points in general. The key issue is how to distinguish suspected outliers in each dimensional subspace.

Therefore, in this proposed method, we project one cell's points of a certain dimension to other dimensions, and compare the points' distribution whether they change or not in the projected dimension. We can find the difference from the distinct features between outlier and normal points. Outliers cannot be clustered with other normal points in all dimensions; they may deviate from other points in different projected dimensions. Nevertheless, normal points that belong to a cluster may keep the near distance each other even in a projected dimensional subspace. We can take the points within the continuous cells as points in one cluster. Consequently, the cell-cluster density of outliers is always lower than the average cell-cluster density of the same cell's points in some other dimensions.

The calculation formula of the cell-cluster ratio against average value is defined in Equation (2), w.r.t. a point $p_{i,j}$ in the $CellNum_{i,k}$ in the $i^{th}$ dimension is projected to the $k^{th}$ dimension.

$$PtValp_{i \to k,j} \Leftarrow CellValp(p_{i,j},k) = \frac{CellNum(p_{k,j}) \times CluLen_j}{\frac{1}{s}\sum_{f=1}^{s} CellNum(p_{k,f}) \times CluLen_f} \quad (2)$$

Where the $CluLen_j$ is the cluster length according to the cell of $p_j$ in $k^{th}$ dimension, $CellNum(p_{k,j})$ is the number of points in the cell where the point $p_j$ is projected in $k^{th}$ dimension from $i^{th}$ dimension, and $s$ is the number of points in the same cell of $p_{i,j}$. The denominator is the average value of the points by the product of $SellNum(p_{k,f})$ and the $CluLen$ in dimension $k$. Here, we compare the points of one cell that $p_{i,j}$ is in with different projected dimensions.

Similar to the relation between $PtVal$ and $CellVal$ in Equation (1), in Equation (2), the point's $PtValp$ gets the value from the $CellValp$ in each projected dimension and $CellValp$ denote the points in the projected cell with proper $PtValp$. After calculation with the loop in projected dimensions, the points with the least total $PtValp$ are more probably to be outliers, too.

For example, taking the Figure 1 to consider the 5 points of the 3$^{rd}$ cell in dimension 2 and to project these 5 points to 5$^{th}$ dimension, each point of $CellValp$ is $CellValp(P_{2,1},5)=0.33$, $SellValp(P_{2,j},5)=2$ $(j=2,3,4)$, and $SellValp(P_{2,5},5)=0.67$.

Most of the time, the points of a cell are no longer clustered into one cell after projecting them to other dimensions. However, normal points still distribute in continuous cells and close to each other. Therefore, the *cell-cluster* is introduced to measure the points' scattering. The outliers are those points far from the normal clustered points or on the edge of huge cell cluster. In this case, their *CellValp* are relative small values.

**Step 3: Statistic information of points with *SI* values**

From the above two steps, each point get $m^2$ times weight values as shown in Equation (4). These weights need to be integrated into one weight value in order to be compared with other points'. Hence, the statistic method is introduced to unite those weight values to one weight value. Here, we use the basic method with summing up all *PtVal* and *PtValp* and being divided by the number of values. The SI values are expressed as reciprocal of the result, to show outliers clearly with highest weight, as shown in Equation (3).

$$SI(p_j) = \frac{2m}{\sum_{i=1}^{m}(PtVal_{i,j}^2 + \frac{1}{m-1} \times \sum_{k=1}^{m-1} PtValp^2(p_{i,j},k))} \quad (3)$$

Where $m$ is the number of dimension, *PtVal* is the value of the point $p_{i,j}$ with Equation (1) in the first projected dimension $i$, *PtValp* is the value of the point $p_{i,j}$ with Equation (2) in second projected dimension $k$, and then $SI(p_j)$ is the statistic information value of $p_j$. The points whose *SI* values are much larger than other points are to be taken as outliers. We count the weights of each point, as shown in Equation (4). The $m^2$ weights guarantee the evaluation of points at a higher accuracy than any previous subspace detection methods.

$$Count(Point\_weight) = m + m \times (m-1) = m^2 \quad (4)$$

Where *Count(Point_weight)* is the total number of point weight values that are a sum of $m$ by the 1$^{st}$ projection and $m \times (m-1)$ by the 2$^{nd}$ projection.

## 3.4 Algorithm

The detailed algorithm is shown in Figure 3 where the dataset contains $n$ points with $m$ dimension, and $K$ is the number of outliers. The algorithm in Figure 3 is written in pseudo R code. The *cn* is an input parameter, and *average cell density* is obtained by calculation. We use the matrix *PointInfo* and the matrix *CellInfo* to record the initial point and cell data information such as point ID, Cell ID and Dimension ID. The PtC*ellid* is original cell id and CellID is used to record the temporary cell ID In projected dimension. The different Cell ID of point are used to build connection between original dimension and the projected dimension. To express the complex relations among points, cells and dimensions, the data structure are introduced below.

*PointInfo[Dimension ID, Point ID]:point in cell ID*

*CellInfo[Dimension ID, Cell ID]: #points in the cell*

With these two data structures, we can find which cell the $p_{i,j}$ is in, and how many points are in the cell. The further information of point and cell in the second projected-dimension is also obtained by inquiring to *PointInfo* and *CellInfo*.

**Figure 3: PCD Algorithm**

| Algorithm: Projected Cell Density |
|---|
| **Input:** data[n,m], cn |
| **Output:** Top K points with largest *SI* value |
| **Begin** |
| 1: Initialize(PointInfo, CellInfo) |
| 2: For i=1 to m |
| 3: d$_i$=n/length(CellInfo[i,]!=0) |
| 4: For j=1 to n |
| 5: Get $PtVal_{i,j}$ and $CellVal(p_{i,j})$ with Equation (1) |
| 6: End j |
| 7: End i |
| 8: For k=1 to 5 |
| 9: Rearrange dimensions in a random order |
| 10: For i=1 to m |
| 11: New Cell=0 |
| 12: For j=1 to n |
| 13: Consider the points in the same cell. |
| 14: Ptcellid= PointInfo[i,j] |
| 15: CellID = PointInfo[i+1, PointInfo[i,]=Ptcellid] |
| 16: CellNum= Count(CellID) +1 |
| 17: Get *CluLen* by counting continuous CellID |
| 18: iff j<n, Get $PtValp_{i+1,j}$ and $CellVal(p_{i,j},i+1)$ with |

```
       Equation(2)
19:      iff j==n, Get CellVal($p_{i,j}$, 1) with Equation(2)
20:     End j
21:   End i
22:  End k
23:  Get SI value for each point with Equation(3)
24:  Output:  top K points with largest SI values from n points
End
```

According to the experiments, some details about *cn* and *average cell density* are to be noticed as follows:

- Before the construction of the cell data space, the parameter *cn* need to be determined.
- For larger *cn* and *average cell density*, it can get better results for points' comparison. Nevertheless, the both values cannot be too small to measure different points.
- The *cn* is inversely proportional to *average cell density* because the total number of points is the product of *cn* and *average cell density*.

Base on above reasons, it is a better choice to set both values strategically to around the square root of the total number of points.

We notice that the points are compared differently in the first dimension loop step and second dimension loop step. In the first step, we compare points with other all points in that dimension. Outliers are found far away or very low density from other points. In the second step, we compare the points in the cell in a dimension. After projected to other dimensions, the normal points still keep near distance, while outliers deviate far from others. Through the twice comparison for targets of different points, the outlier are separated from others successfully.

## 4. Evaluation

The experiments and the evaluations on two-dimensional data are given in Section 4.1 by comparing our proposal with ever-proposed methods, and those on high dimensional data are given in Section 4.2.

All the experiments were performed on MacBook Pro with 2.53GHz Intel core 2 CPU and 4G memory. The proposed algorithm was implemented with R language on Mac OS.

We usually need to find some test data such as experimental data or practical data to evaluate the proposed method. However, it is difficult to define which points are exact outliers and how the noise data affect the result in the real datasets, especially in the case of more than 100 dimensional dataset. Therefore, we decided to generate a series of data to evaluate our proposal. The generated datasets include outliers and normal points according to Hawkins' definition.

The rules of generating high dimensional artificial dataset in this paper are listed below:

- Normal points' and outliers are distributed in the same region. The range of both data is overlapped each other.
- Normal points' and outliers belong to different distributions, respectively.
- Outliers are far less than normal points. The number of outliers is usually less than 5-10% of total points.

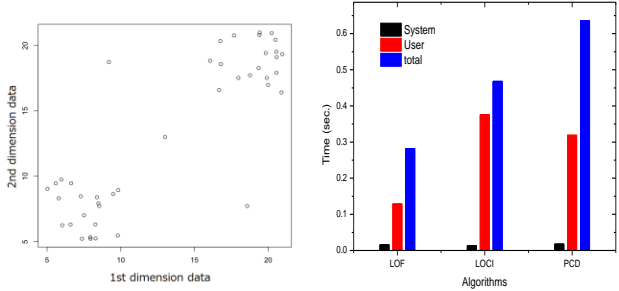

**Figure 4: Data Distribution &Algorithms' Processing Time in Two-Dimensional Experiments**

- Outlier should not be observed in low projected-dimensional space.

The generated dataset with high dimensions is clarified at length in Section 4.2. The parameters cited in the dataset is shown in Table 2.

As for the evaluation, F-measure, precision and recall are used to compare the proposal with LOF and LOCI methods.

### 4.1 Two-Dimensional Data Experiment

In the two-dimensional experiment, we generate 43 points, which include 40 normal points and 3 outliers as shown in Figure 4. According to this figure, the half normal points are randomly distributed in a region from 5 to 10 in both dimensions, and the other half are randomly distributed in a region from 16 to 21 in both dimensions. The 3 points are as outliers placed in the middle of both regions since they do not belong to any clusters of normal points.

The purpose of designing two-dimension data distribution is to verify our proposal to be effective or not even in low dimensional space. As shown in Figure 4, the center outlier is far from normal points in each dimension. So this outlier can be detected in the step 1 as described in Section 3.3. On the other hand, because the other two outliers mixed with normal points in each dimension, these outliers can be detected only in the projected dimension in the step 2 as described in Section 3.3.

To evaluate the experimental results, we compare our proposal with the well known algorithm LOF[2] and LOCI[6]. In the two-dimensional data, the three algorithms achieve the perfect results with 100% precision and 100% recall. The processing time of each is compared, as shown in Figure 4.

According to Figure 4, the LOF is the fastest algorithm and LOCI's user time takes more than that of our proposal. However, the total time is shorter than that of our proposal. Hence, we can conclude that our proposed method is also able to detect the outliers in low dimensional space, and accuracy is as same as the conventional methods.

### 4.2 High Dimensional data Experiment

In the high dimensional experiments, we generate eight data sets with 10-10000 dimensions, where the data size is 500 or 1000. In order to generate suitable experimental datasets, we refer to Hawkins'outlier definition [1] and Kriegel's dataset model [4]. The experimental data set includes normal data with Gaussian mixture model consisting of five equally weighted normal distributions, whose means is randomly ranged from 20 to 80 with random variance values ranged from 5 to 20. Each experimental

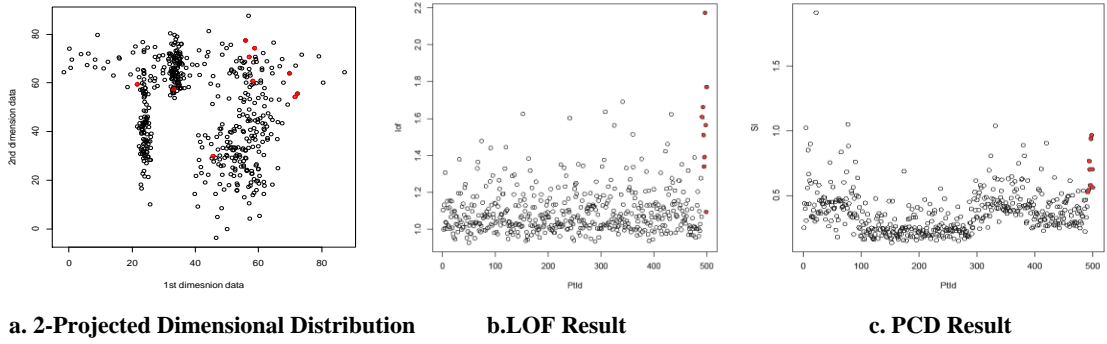

**a. 2-Projected Dimensional Distribution    b.LOF Result    c. PCD Result**

**Figure 6: Example of dataset 500 Points with 10 Dimensions**

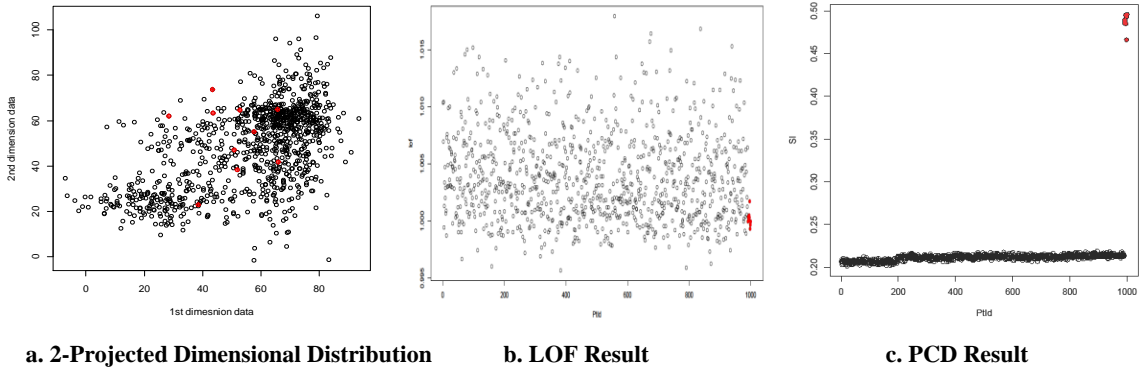

**a. 2-Projected Dimensional Distribution    b. LOF Result    c. PCD Result**

**Figure 7:  Example of dataset 1000 Points with 10000 Dimensions**

dataset includes 10 outliers with random distribution in all dimensional space, whose range is from 20 to 80, exactly inside the range of the normal points. The detail parameters of generated datasets are listed in Table 2.

**Table 2:  Parameters of Artificial Dataset**

| Dataset No. | Dimension number | Points number | Normal points (Normal distribution) | Outliers Random |
|---|---|---|---|---|
| 1 | 10 | 500 | µ(20-80), σ(10-20) | (20-100) |
| 2 | 100 | 500 | µ(20-80), σ(10-20) | (20-100) |
| 3 | 100 | 1000 | µ(20-80), σ(10-20) | (20-100) |
| 4 | 500 | 500 | µ(20-80), σ(10-20) | (20-100) |
| 5 | 500 | 1000 | µ(20-80), σ(10-20) | (20-100) |
| 6 | 1000 | 500 | µ(20-80), σ(10-20) | (20-100) |
| 7 | 1000 | 1000 | µ(20-80), σ(10-20) | (20-100) |
| 8 | 10000 | 1000 | µ(20-80), σ(10-20) | (20-100) |

Our outliers' generation is more restricted than that of Kriegel's dataset model for two reasons. First, while Kriegel's outliers are uniform distribution that easily to be detected, ours are randomly distributed and hidden in normal points. Second, Kriegel's outliers are distributed in a whole range that why some of them are far from normal points in low dimensional space, while all of our outliers are distributed inside the normal points' range that make the outlier detection difficult in low dimensional space.

The LOF and LOCI algorithms are chosen initially as to be compared with our proposed PCD. Nevertheless, LOCI experiment results are poor in almost all experiments. The best result with LOCI is in the case of 100 points and 1000-dimension dataset, where the precision and recall are 20% and 25%, respectively. In other datasets, the LOCI got either no outliers or false outliers. Hence, LOCI is not an effective method to detect these inner outliers in high dimensional space. Based on this reason, we only select LOF algorithm to compare with our proposed PCD in this section.

As for the initial parameters, *cn* is set to 25 and 35 for 500 points and 1000 points datasets, respectively, in PCD. The parameter 10 is set for *MinPts* as a k-nearest neighbor value in LOF.

In Figure 6, the result of LOF and the result of PCD are shown while dataset 8 and the results of LOF and PCD are shown in Figure 7. In both figures, the outliers are labeled in red x while the normal points are in the black circle.

The outliers are assigned inside the region of normal points. Therefore, in Figure 6(a), the red outliers cannot be found in projected subspace just by observing the distributions. Even after the processing with LOF and PCD, only small part of outliers can be detected at low precision. The inner outliers are difficult to be detected in low dimensional space since the noisy points are more likely anomaly than they are. Figure 7(a) looks similar to Figure 6(a), but the number of dimensional is higher. In this case, the result of LOF as shown in Figure 7(b) shows that it performs poorly in such high dimensional dataset, while our proposed PCD shows the perfect result even in this high dimensional space.

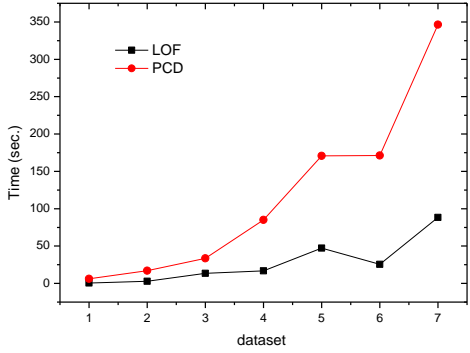

**Figure 8: Processing Time in High Dimensional Experiments**

All the experimental results are summarized in Table 3. In the experiments, the most suitable thresholds in both algorithms were given as also shown in Table 3.

**Table 3: Algorithms' Result in High Dimensional Experiments**

| ID | Dimension | Point | LOF | | | | PSD | | | |
|---|---|---|---|---|---|---|---|---|---|---|
| | | | Threshold | Precision | Recall | F-measure | Threshold | Precision | Recall | F-measure |
| 1 | 10 | 500 | 1.5100 | 0.7 | 0.5000 | 0.5833 | 0.7 | 0.7000 | 0.0496 | 0.0927152 |
| 2 | 100 | 500 | 1.2200 | 0.8 | 0.8889 | 0.8421 | 0.4 | 0.8000 | 1.0000 | 0.8888889 |
| 3 | 100 | 1000 | 1.2100 | 1.0 | 0.9091 | 0.9524 | 0.34 | 1.0000 | 1.0000 | 1 |
| 4 | 500 | 500 | 1.1200 | 0.1 | 1.0000 | 0.1818 | 0.49 | 1.0000 | 1.0000 | 1 |
| 5 | 500 | 1000 | 1.2700 | 0.1 | 1.0000 | 0.1818 | 0.4 | 1.0000 | 1.0000 | 1 |
| 6 | 1000 | 500 | 0.9964 | 0.1 | 1.0000 | 0.1818 | 0.5 | 1.0000 | 1.0000 | 1 |
| 7 | 1000 | 1000 | 1.0000 | 1.0 | 0.0105 | 0.0208 | 0.4 | 1.0000 | 1.0000 | 1 |
| 8 | 10000 | 1000 | 0.9993 | 0.8 | 0.0976 | 0.1740 | 0.4 | 1.0000 | 1.0000 | 1 |

The experiment results show that the LOF algorithm produces the poor result in all datasets. Though the LOF perform better in 10-dimensional dataset than PCD, the F-measure value is still around 55%. In fact, both algorithms perform poorly in 10-dimensional space. However, our proposal gives perfect results when the dimension increases. In the experiment of dataset around 100 dimensions, the F-measure is close to 100%. After that, all results are perfect with 100% F-measure. LOF algorithm can find outliers around 100-dimension, although the precision is only 95.24%. When the dimension increases to 500, LOF hardly finds the outliers. In contrast, our proposed algorithm performs well when dimension increases very high. When dimension is up to 100, the precision and recall are almost 100%. For both algorithms, their precisions raise when the number of points increase. As the conclusion, our proposed algorithm performs better than LOF in high dimensional space, not only on evaluation with the precision and the recall, but also more stable when the dimension increases.

Regarding to Figure 8, the LOF algorithm is faster than our algorithm, because it only considers the distance between points regardless number of the dimensions; while our proposal considers the both number of dimension and number of points.

Therefore, our proposal requires more time in projected dimension computation. The result also shows the both algorithms process more time with the growing of points and dimensions.

The experiment results are shown with chart in Figure 9. It clear shows that PCD is dominated advantage on LOF except first dataset test. LOF performs well in relative high dimensional dataset. However, LOF performs worse when dimension increase extreme high in the later five datasets experiments.

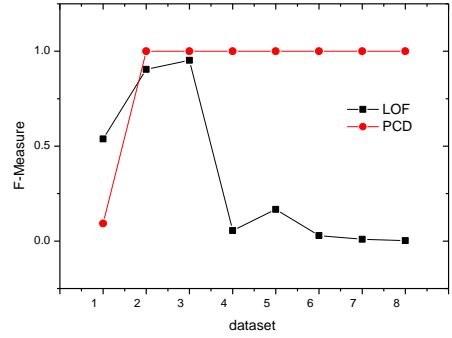

**Figure 9: Results with F-measure in High Dimensional Experiments**

## 5. CONCLUSION

In this paper, we propose a new algorithm for outlier detection in high dimensional space. The proposed algorithm not only works smoothly in low dimensional space, but also runs effectively in high dimensional space. In the experiments, our algorithm has performed better in the high dimensional datasets compared with LOF approach. In other words, the "curse of dimensionality" problem has been solved to some extent in our proposed algorithm. Moreover, our proposed solution has provided a general framework to solve the similar high dimension problems. Hence, many existing methods or new methods can be applied under this framework.

Our proposed algorithm is a general approach to find anomaly data in large-scale database. Therefore, it may apply to multimedia field, e.g. Discovering some remarkable images from billions of images and denote the new annotation to these images. This is a basic task for image annotation. The PCD method supplies a thorough detection method in high dimensional dataset, which would find the trivial difference among images that contain a large number of attributes.

To be noted that, our proposed algorithm is composed of two steps that detect outliers in each dimension and each projected dimension. We apply the equal weight to these two steps. However, the optimal weighting values to these attributes needs to be clarified as one of our future works.

One of the crucial issues is the processing time of the large cycle calculation because the calculations in the projected dimension are independent each other. Consequently, the parallel processing could bring some benefits to our proposal to reduce the processing time. Furthermore, it is also another critical issue to find the possibility to apply our proposed algorithm to other methods like clustering, classification, etc. For the reason the "curse of dimensionality" is still a necessary issue need to be concerned even in these methods.